\documentclass[letterpaper, 10 pt, conference]{ieeeconf}  

\IEEEoverridecommandlockouts     

\usepackage{graphics} 
\usepackage{epsfig} 
\usepackage{tabu}
\usepackage{multirow}

\title{\LARGE \bf
A Lobster-inspired Robotic Glove for Hand Rehabilitation
}

\author{Yaohui Chen$^{}$, Sing Le$^{}$, Qiao Chu Tan$^{}$, Oscar Lau$^{}$, Fang Wan$^{}$, Chaoyang Song$^{*}$, \textit{Member, IEEE}
\thanks{*Correspondence author}
\thanks{Y. Chen, L. Sing, Q. Tan, and O. Lau are students with the Faculty of Engineering, Monash University, Clayton, VIC, 3168, Australia (email: {\small yaohui.chen@monash.edu}; {\small \{sle11, qctan1, olau2\}@student.monash.edu}.)
\newline \indent F. Wan was with Nanyang Technological University, Singapore and now is an independent researcher. (email:  {\small sophie.fwan@gmail.com})
\newline \indent C. Song is Lecturer with the Faculty of Engineering, Monash University, Clayton, VIC, 3168, Australia (phone: +61-03-99052994, email: 
{\small chaoyang.song@monash.edu}).}%
}

\begin{document}

\maketitle
\thispagestyle{empty}
\pagestyle{empty}

\begin{abstract}

This paper presents preliminary results of the design, development, and evaluation of a hand rehabilitation glove fabricated using lobster-inspired hybrid design with rigid and soft components for actuation. Inspired by the bending abdomen of lobsters, hybrid actuators are built with serially jointed rigid shells actuated by pressurized soft chambers inside to generate bending motions. Such bio-inspiration absorbs features from the classical rigid-bodied robotics with precisely-defined motion generation, as well as the emerging soft robotics with light-weight, physically safe, and adaptive actuation. The fabrication procedure is described, followed by experiments to mechanically characterize these actuators. Finally, an open-palm glove design integrated with these hybrid actuators are presented for a qualitative case study. A hand rehabilitation system is developed by learning patterns of the sEMG signals from the user’s forearm to train the assistive glove for hand rehabilitation exercises.

\end{abstract}

\section{INTRODUCTION}

For patients with surgically repaired fingers after traumatic injury, hand rehabilitation is needed to help them regain finger strength and mobility. Repetitive exercises are suggested to move the finger joints through the motion range \cite{Rosenstein2008EffectsStroke}. For example, one of the most common exercises is to force the fingers to move from an open state to a closed one \cite{Polygerinos2013TowardsRehabilitation}. Instead of attending physiotherapist-assisted programs, a robotic system that assists patients to carry out exercises on their own would be a more convenient and economic choice. However, the challenge remains to build such a system within an affordable trade-off between cost and benefit. 

In the past four decades, research and development in glove-based systems presented a shifting focus from sensing \cite{Dipietro2008AApplications} to actuation \cite{Heo2012CurrentEngineering} in engineering applications ranging from gesture recognition to bio-medical science. Many robotic gloves are designed and built with rigidly powered transmission mechanisms for active finger actuation. Such rigid designs like Rutgers Master II \cite{Bouzit2002TheGlove} as well as many others \cite{Lum2012RoboticStroke, Takahashi2005AStroke, Kawasaki2007DevelopmentControl} usually present powerful and precise motion generation. However, they suffer drawbacks in mechanism compliance and adaptive motions conforming to human fingers \cite{Heo2012CurrentEngineering}. Engineering applications of such rigid glove may seek more practical applications in power gloves or hand exoskeletons augmenting human grasping capabilities, instead of personal use for hand rehabilitation that requires more delicate design concerns in safety, adaptation, and portability. 

\begin{figure}[tbp]
    \centering\includegraphics[width=1\linewidth]{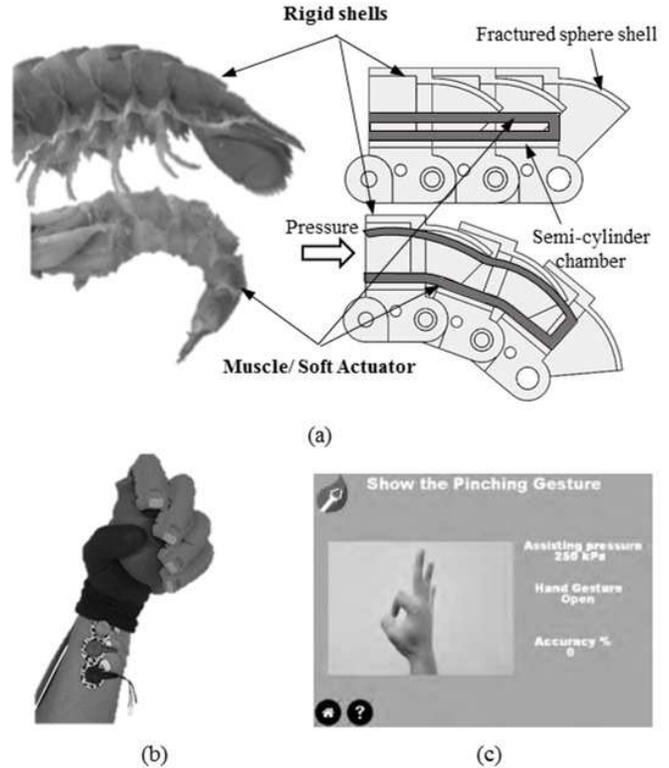}
    \caption{A lobster-inspired hybrid robotic glove for hand rehabilitation: (a) design inspiration from the lobster abdomen and the inspired hybrid actuator; (b) integration of the hybrid actuator into an assistive robotic glove; (c) a user interface for hand rehabilitation exercise of the rehabilitation system.}
    \label{fig:DesignInspiration}
\end{figure}

Recent development in soft robotics brings a new dimension to the material property to the design of robotic gloves, especially those for hand rehabilitation \cite{Meng2015AStructure,  Noritsugu2004WearableMuscle, Connelly2010AStroke}. Such soft robot designs present with promising potentials to be a novel solution that is safe, adaptive and wearable. However, soft actuators have some inherent disadvantages like low repeatability in production with the complex fabrication process \cite{Agarwal2016StretchableDevices} and vulnerability to ruptures upon inflation \cite{Paez2016DesignReinforcement}. To avoid these issues, some new designs for soft actuators have been proposed to simplify the fabrication chain or constrain the radial expansion of soft actuators \cite{Marchese2015ARobots, Roche2014AMaterial,  Polygerinos2015ModelingActuators}. Recently, outer casing shell with cuts \cite{Agarwal2016StretchableDevices, Memarian2015ModellingMuscles} and corrugated outer fabric \cite{Yap2016AApplication}, which are more repeatable in fabrication, are used in to provide a more robust constraint to the undesired inflation, but local buckling still cannot be completely avoided. Origami is used as an additional protection layer \cite{Paez2016DesignReinforcement} throughout the bending range in one research. To improve the design of robotic gloves for hand rehabilitation, there is a strong motivation towards developing new actuators with a simpler fabrication process and stronger protection during the whole motion range. 

In this paper, we present a novel design of robotic glove for hand rehabilitation which utilizes a new lobster-inspired actuator design with both rigid and soft components (Fig. \ref{fig:DesignInspiration}(a)). The modular rigid shells constrain the radial inflation of the soft chamber inside and provide a full protection throughout the range of bending motion. In the meanwhile, the soft chamber provides necessary actuation through interaction with the rigid shells. The rigid shells can be 3D printed and the soft chamber takes a simple rectangular cross-section fabricated through a simple molding process, which minimizes the manual work and significantly simplifies the fabrication process. This kind of novel hybrid actuators are then assembled into a robotic glove for hand rehabilitation (see Fig. \ref{fig:DesignInspiration}(b)), demonstrating the ease of configuring this hybrid actuator to complex robotic devices. A game-based rehabilitation program (see Fig. \ref{fig:DesignInspiration}(c)) is developed to learn from the sEMG signal patterns from the forearm of a healthy hand to train the patient's dysfunctional hand for finger motor reconstruction. 

In section 2, details of actuator design, fabrication and characterization are presented. Especially, two versions of hybrid actuators were developed to illustrate how shell design and assembling strategy influence the actuators' mechanical performances. In section 3, a robotic glove design is proposed using the first version of hybrid actuators (actuator V1), validating the rationality to adopt hybrid actuators into wearable robotic devices. A rehabilitation system was developed using the second version of hybrid actuators (actuator V2) with more robust characteristics, and details of the sEMG controlled, portable solution with a game interface for hand rehabilitation protocols are presented in section 4. Conclusions are enclosed in the final section, which outlines agenda for future development of the system for hand rehabilitation.  

\section{ACTUATOR DESIGN \& CHARACTERIZATION}

\subsection{Actuator Design and Fabrication}

A conceptual hybrid bending actuator is proposed based on the bending motion from the lobster abdomen, including its rigid exoskeleton and soft muscle as in Fig. \ref{fig:DesignInspiration}(a). To bio-mimic this structure, the rigid shells of the hybrid actuator takes a modular design with a semi-cylinder chamber combined with a fraction of a spherical shell. Similar to the lobster's rigid exoskeleton, once jointed, the next shell segment's fractured spherical shell can be folded inside the previous shell segment's semi-cylinder chamber, forming a deployable mechanism with bending motion. The actuation is produced by the soft chamber placed inside the rigid shells under pressurization. When inflated, the soft chamber pushes against the inner walls of the rigid shells, especially the edge of the fractured spherical shell, which will then produce a bending motion through the next shell segment. 

\begin{figure}[htbp]
    \centering\includegraphics[width=1\linewidth]{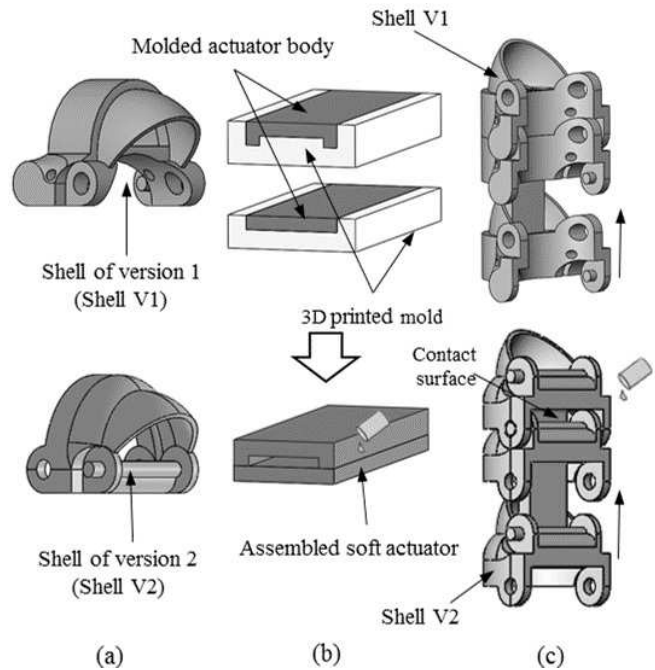}
    \caption{Fabrication process using 3D printed molds, silicone and 3D printed shells. (a) Two version of shell designs. (b) Two-step molding process for the soft actuator. (c) Two assembling strategies to assemble the soft actuator with rigid shells.}
    \label{fig:Fabrication}
\end{figure}

Two designs of the rigid shells are shown in Fig. \ref{fig:Fabrication}(a). For version 1 of the shell design, the semi-cylinder chamber has a length of 8mm and a radius of 8mm. The total width of the shell V1 was 22mm and height was 14mm. Version 2 of the rigid shell were designed with similar geometry as version 1 in the length and radius of the semi-cylinder. However, the total width and height were reduced to 18mm and 12mm with a simplified joint design to reduce friction on the joint axis and increase the internal space in the fractured sphere section. The angles of the fractured sphere were both measured at ${45^\circ}$ about the horizontal plane. Another difference is the different 3D printers used for both shell versions. V1 shell was printed using an FMD based printer by 3D System with a printing resolution of 0.1mm. V2 shell was printed with an SLA based printer by formlab with a much-improved printing resolution of 0.025mm. As a result, although the overall geometry is reduced in the V2 shell, the volume inside the fractured sphere section is actually increased to allow more effective and efficient bending motion.

With the enclosure of rigid shells, the soft actuator inside is free to take any shape as long as it fits inside the rigid shells. In our prototype, a simple rectangular chamber of 10mm by 3.5mm was fabricated with a wall thickness of 1.5 mm. As shown in Fig. \ref{fig:Fabrication}(b), a simple two-step molding process was adopted. Equal parts of Ecoflex 00-30 and Dragon Skin 30 (Smooth-On, Inc.), were mixed to produce a desirable elasticity and strength of the soft actuator. First, the upper layer mold was used to create the air chamber and roof, while the bottom mold to form the base. Then, the two parts were sealed together using a thin layer of the same mixture. The open end of the soft actuator was fitted with a tube connector and sealed using a cable tie and Silpoxy silicone glue. 

To assemble the hybrid actuators, two assembling strategies were adopted as shown in Fig. \ref{fig:Fabrication}(c). For actuator V1, the soft actuator was fed through the shells V1 and fixed at the tip segments. While for actuator V2, an additional step after the same procedure was added to glue the contact surfaces between the soft actuator and rigid shells to ensure a uniform stretch along the length direction upon inflation.  

\subsection{Actuator Characterization}

For actuator V1, a configuration with 10 rigid shells jointed in serial was cantilevered on a mounting for the experimental testing. By removing the number of rigid shells at the free end of the actuator and cutting the length of the soft chamber inside, 9 different configurations (10 segments to 2 segments) of the hybrid actuator were characterized with regard to its bending motion and force output. Using the same method, 6 different configurations (12 segments to 7 segments) were tested with actuator V2.

\begin{figure}[htbp]
    \centering\includegraphics[width=1\linewidth]{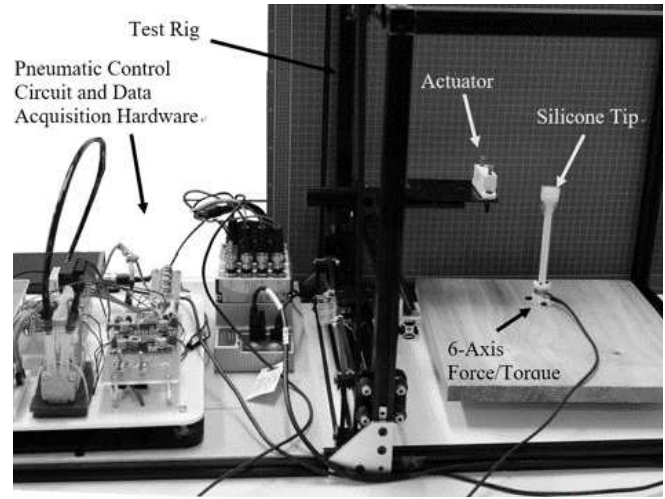}
    \caption{Test platform with pneumatic controllers to characterize the hybrid actuator.}
    \label{fig:Platform}
\end{figure}

A characterization platform was built containing a 6-axis force/torque sensor mounted onto a three-axis adjustable aluminum frame to measure the force at the tip of the hybrid actuators (see Fig. \ref{fig:Platform}). A graphical interface (GUI) written in LabView was developed to control the electronics. The pressure inside the soft actuator was read through a microcontroller (Arduino Mega) and presented to the user via the screen. In the force test, the proximal tip segments of the actuators were tightly clamped to emulate a fixed boundary condition. A 3D printed post was mounted on top of the force/torque sensor to bring contact to the bottom of distal tip segments. A constraining platform was placed on top of hybrid actuators to keep them in zero bending angles under actuation. In the bending test, the distal tips of hybrid actuators were free to bend due to input air pressure. A high definition camera was mounted on a tripod to record the bending trajectory from the side view. Post-processing of motion sequence was conducted in Adobe Illustrator.

\begin{figure}[htbp]
    \centering\includegraphics[width=1\linewidth]{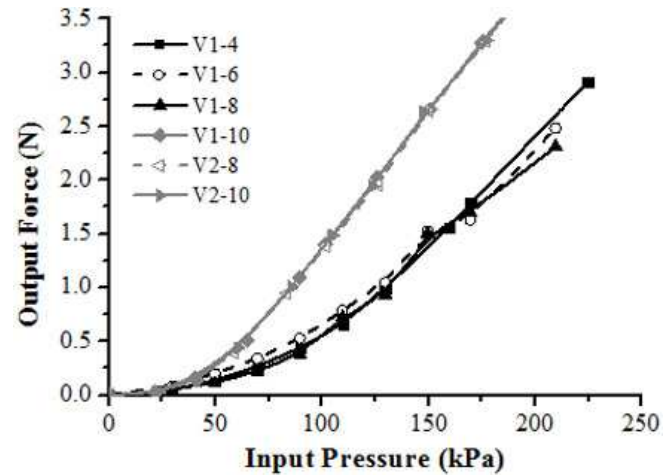}
    \caption{Force result of two versions of hybrid actuators with different segments}
    \label{fig:Force}
\end{figure}

Results of the force test were presented in Fig. \ref{fig:Force} for both V1 and V2 hybrid actuators. The pressure inside the soft chamber was gradually increased up to 230kPa for actuator V1 and 190kPa for hybrid actuator V2, and the exerting force at the distal tip was recorded. The test was repeated 3 times for each configuration for accuracy and consistency. A positive correlation was observed between the applied pressure and output force. For hybrid actuator V1, the tip force was measured at about 2.5N under 210kPa with a different number of segments. For hybrid actuator V2, the measured tip force was increased to around 3.5N under just 180kPa. One possible reason for the improvement in force output in actuator V2 could be the assembling strategy adopted, which glued the contact surfaces between the soft actuator and rigid segments and constrained the stretch in length direction upon pressuring. A larger internal space in the fractured sphere section and better resolution of the 3D-printed rigid shells in actuator V2 could also contribute to the increase of tip force and will be further investigated. It was noted that segment numbers had a small influence on tip force output for both versions of actuators.

\begin{figure}[htbp]
    \centering\includegraphics[width=1\linewidth]{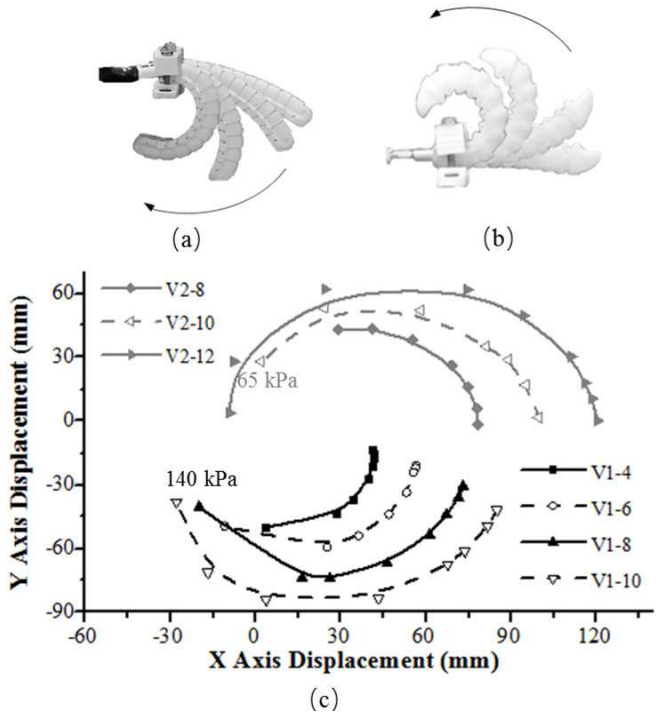}
    \caption{Bending test of the hybrid actuator: (a) Bending motion sequence of actuator V1. (b) Bending motion sequence of actuator V2. (b) Bending trajectories of two versions of hybrid actuators with different configurations.}
    \label{fig:Bending}
\end{figure}

In the bending test, different configurations of actuator V1 were clamped with the flat surface downwards. Their tip segments hung down a little before pressuring due to gravity, and bent clockwise with increasing pressure inside the soft actuator as shown in Fig. \ref{fig:Bending}(a). On the contrary, the flat surface of actuator V2 was clamped upwards.  Gravity-generated hang was avoided due to geometric constraints, and they bent anticlockwise upon inflation (see Fig. \ref{fig:Bending}(b)). The pressure was gradually increased with increments of about 20kPa for actuator V1 while about 5kPa for actuator V2, and results of bending trajectories were presented in Fig. \ref{fig:Bending}(c). The position where the proximal tip segment was clamped was recorded as the origin for all configurations. It is clearly that bending capacity of hybrid actuators increased with increasing segment numbers. One possible reason for the performance difference of two versions could be that additional stretch in length direction occurred in actuator V1 during pressuring, as they lacked constraints between contact surfaces of the soft actuator and rigid segments. Geometric differences in fractured sphere section could also influence the results that requires further investigation.  

\section{Soft Robotic Glove}
In this study, the proposed hybrid actuator was further integrated into the rehabilitation glove design. A portable pneumatic and control system was developed along with a hand rehabilitation game to assist the rehabilitation process. Especially, we explore a strategy using the unimpaired hand to control the activation of the robotic glove on the impaired hand, and surface electromyography (sEMG) signals are collected for gesture characterization. In this study, only one muscle signal was detected and different muscle motions were classified by K-nearest neighbor (KNN) algorithm to simplify the application scenario for at-home rehabilitation.

\subsection{Quick Soft Robotic Glove Design}

\begin{figure}[htbp]
    \centering\includegraphics[width=1\linewidth]{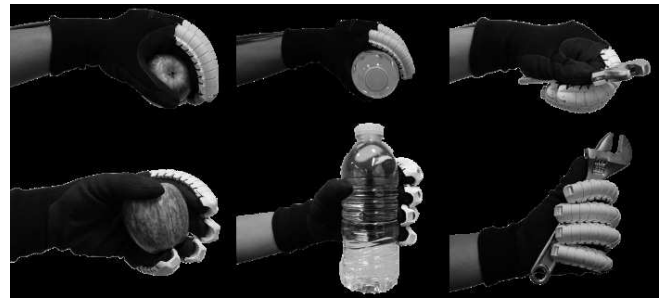}
    \caption{Manipulation tasks of the robotic glove made from actuator V1 conforming to objects with different shapes, sizes, textures and weights.}
    \label{fig:GraspingTask}
\end{figure}

To quickly evaluate the ease of adopting this novel hybrid actuator design into wearable robotic systems, actuator V1 was integrated into an assistive robotic glove as shown in Fig. \ref{fig:GraspingTask}.

With the modular design of the hybrid actuator, a different number of segments could be sewed onto a flexible nylon glove (Cyclone Gel Glove). It ensures a quick, easy and safe fitting to the human hand to provide conformability in maximum. The sliding issue with such rigid shell design is minimized by shifting the joint axis so that it aligns with the human finger joint axis. For this prototype, the middle finger used 10 segments, the ring and index fingers used 9, and pinky used 7. The thumb was currently excluded from the glove. The total weight of the whole glove was measured approximately 150 grams, including tube fittings at the end of the soft actuators. Additionally, each finger could be actuated individually with its own hybrid actuator. These actuators were placed on top of the glove to guide the finger joints in desired motions under actuation.

To evaluate the ability of the robotic glove in assistive grasping, three grasping tasks were conducted with a participant wearing it. The objects examined for grasping were of different shapes, sizes, textures and weights (an apple, a bottle of water and a pincer). The increased pressure inside the hybrid actuators generated a bending motion of the glove to form a closed fist, and an output force could be exerted when in contacting with external objects. The internal pressure was kept at 250kPa to fulfill the three grasping tasks as shown in Fig. \ref{fig:GraspingTask}, validating the potential of adopting this hybrid design into assistive glove devices.   

\subsection{sEMG Classification}

\begin{figure}[htbp]
    \centering\includegraphics[width=1\linewidth]{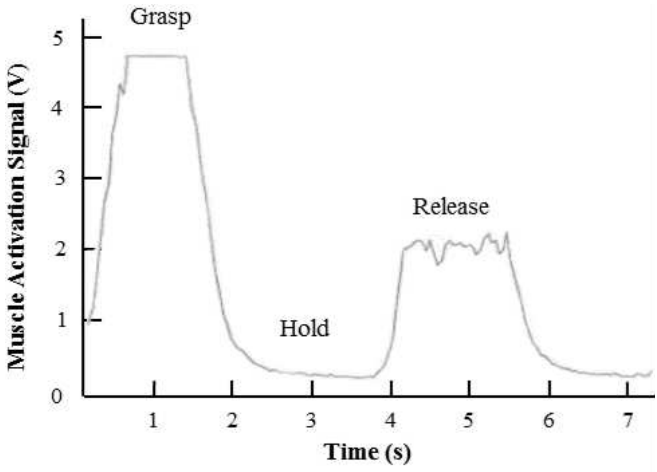}
    \caption{The EMG signal collected from FDP muscles during a grasp-release gesture.}
    \label{fig:EMGSignals}
\end{figure}

Through measuring the electrical activity of muscles on the forearm by detecting the sEMG, user intent of finger flexion and extension can be classified. Single channel electrodes were placed at the Flexor Digitorum Communis (EDC) to detect the muscle signal. The circuit implementation of the sEMG sensor produced a full rectified wave output signal that was used for processing as shown in Fig. \ref{fig:EMGSignals}. The data acquisition system was specially designed to pick up the sEMG signals from the user's forearm, and the signals were processed through a microprocessor (Arduino) which provides real-time graphical data for comparing the signals of different gestures. 

The K-nearest neighbor (KNN) is a non-parametric method used for classification by recognizing the most similar labeled training samples, and was used in this part for the classification of sEMG signals. This algorithm operation is based on comparing an unknown input vector with training samples in the system, and searching the space for the K training records that are nearest to the input vector as the new record neighbors. Five features were extracted for the algorithm calculation including integrated EMG (IEMG), mean absolute value (MAV), simple square integral (SSI), maximum EMG signal (MAX) and waveform length (WL) as listed in Table. \ref{tab:Features}. The K-nearest neighbors can be determined by calculating K minimal Euclidean distances in the system defined as 

\begin{equation}
    \label{eq:distance}
    distance=\sqrt{\sum_{i=K}^5 {(x_{input,i}-x_{sample,i})}^{2}}.
\end{equation}

\begin{table}
    \newcommand{\tabincell}[2]{\begin{tabular}{@{}#1@{}}#2/end{tabular}}
    \centering
    \caption{Features extracted for KNN algorithm}
    \begin{tabular}{m{0.5cm}<{\centering} m{3.5cm}<{\centering} m{3cm}<{\centering}}
    \hline
    \textbf{Feature} & \textbf{Definition} & \textbf{Expression}\\
    \hline
    IEMG & Area under the signal curve & $x_{1}=\sum_1^N V_{i}$ \\
    MAV & Average value of the amplitudes & $x_{2}=\frac{1}{N}\sum_{i=1}^N |V_{i}|$ \\
    SSI & Summation of squared absolute amplitude values & $x_{3}=\sum_{i=1}^N {|V_{i}|}^{2}$\\
    MAX & Maximum signal amplitude & $x_{4}=max(V_{n})$\\
    WL & Cumulative length of the waveform & $x_{5}=\sum_{i=1}^{N-1} |t_{i+1}-t_{i}|$\\
    \hline
    \end{tabular}
    \label{tab:Features}
\end{table}

The method of recognition was generalized in python to obtain numerical values of the classier. The user conducted 34 grasp-release gestures to produce 34 training samples for both grasping and releasing gestures. Then, 16 grasp-release gestures were conducted for validation of the accuracy, which was defined as the ratio of the number of correctly classified sEMG signals to that of total training samples. Influence of different K values was also investigated by modifying the number of nearest neighbors. The efficiency of the system could be determined by calculating the time it took to execute the algorithm. The results concluded in Table \ref{tab:Accuracy} demonstrated the feasibility of this sEMG characterization method with only one muscle signal with good accuracy for real-time hand rehabilitation.

\begin{table}
    \newcommand{\tabincell}[2]{\begin{tabular}{@{}#1@{}}#2/end{tabular}}
    \centering
    \caption{Recognition accuracy and CPU time with different K values}
    \begin{tabular}{m{1cm}<{\centering} m{2cm}<{\centering} m{2cm}<{\centering}}
    \hline
    \textbf{} & \textbf{CPU time (s)} & \textbf{Accuracy (\%)}\\
    \hline
    K=1 & 0.007007 & 79.6 \\
    K=3 & 0.007048 & 81.6 \\
    K=5 & 0.007063 & 85.2\\
    K=7 & 0.00713 & 85.4\\
    K=9 & 0.00713 & 84.9\\
    \hline
    \end{tabular}
    \label{tab:Accuracy}
\end{table}

\subsection{Hand Rehabilitation System}

\begin{figure}[htbp]
    \centering\includegraphics[width=1\linewidth]{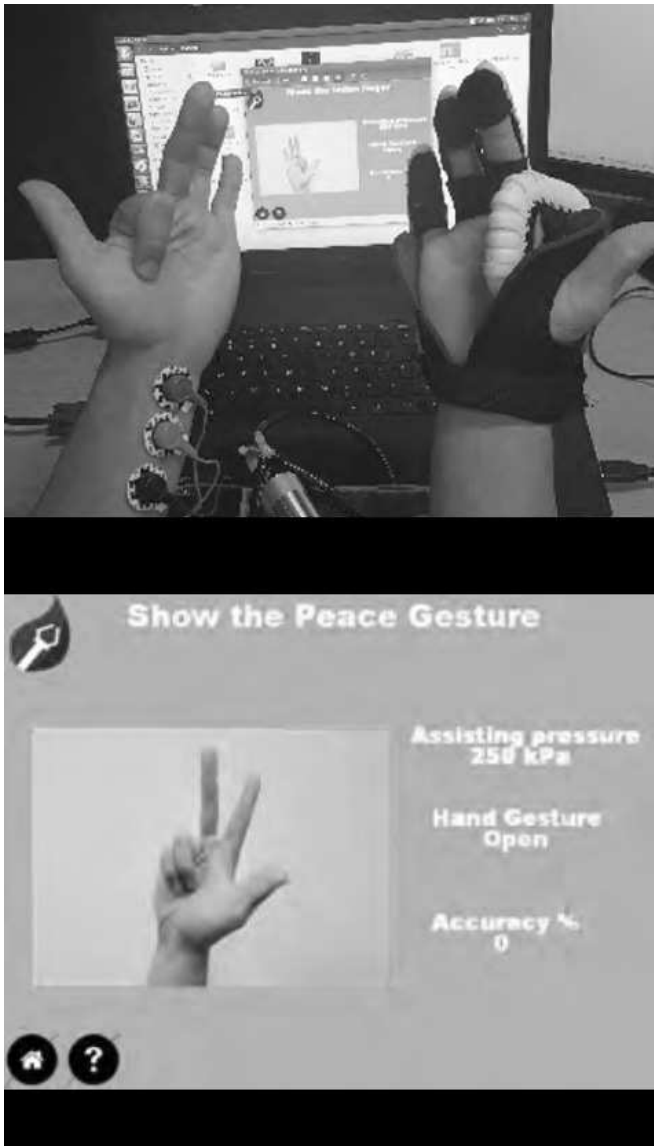}
    \caption{The hand rehabilitation system developed in this study. (a) Implementation of the overall system including the robotic glove made of actuator V2, control system, and a game interface. (b) Instruction displayed on the interface during rehabilitation process.}
    \label{Rehabilitation}
\end{figure}

A hand rehabilitation system was developed with the robotic glove proposed earlier, while actuator V2 was adopted due to its improved performance. A portable system was designed and a game interface was introduced to create an interesting and effective rehabilitation process as shown in Fig. \ref{Rehabilitation}(a).

The user was advised to wear the robotic glove on his impaired hand, and put electrodes on their unimpaired hand. Then, the user could engage with this system by following the instructions displayed on the graphic interface (Fig. \ref{Rehabilitation}(b)) with the unimpaired hand. The electrodes would measure the muscle activity on the forearm and the signals would be classified using KNN algorithm. As the user performed the desired movement, the hand with the robotic glove would be actuated by the pneumatic system, and assistive motion would be provided on the finger that was instructed by the game interface.  

\section{Conclusion and Future Work}
\label{S:4}
This paper presents the preliminary study of a hybrid actuator design inspired by the rigid shells and soft muscles in lobsters, or crustaceans in general. The design complexity is much reduced for the soft actuator in this hybrid design, which can essentially take almost any form as long as it fits the internal geometry of the rigid shells. On the other hand, the rigid shells provide a full protection to the soft actuators throughout the range of motion. Two versions of rigid shells were designed, and mechanical performance of two version of hybrid actuators was experimentally characterized, demonstrating the ability to replicate finger movements for active assistance. Improvements in shell design and assembling improved the force output and bending capacity, which will be further investigated. Moreover, a quick robotic glove design demonstrated the ease to adopting this novel design into assistive devices. A rehabilitation system was developed for at-home rehabilitation. Only one sEMG signal was collected in this system and different gestures were classified by KNN algorithm with good accuracy and efficiency for home use.   

The 3D printed glove prototype exhibit a lightweight design of 150 grams. Such hybrid actuator demonstrates the ability to be customized to a wide range of finger sizes with comfort on the patient’s hand. Although the hybrid actuator was designed specifically for hand rehabilitation in this paper, the same method of actuation can be followed to produce other hybrid actuators that are suitable for rehabilitating patients in other areas such as the elbow or knee joint. The exoskeleton and soft actuators can be modified to suit the purposes of different rehabilitative systems. The hybrid glove is intended to combine assistance with activities of daily living and at home rehabilitation for individuals. The overall system can be compacted into a portable waist belt pack or backpack that offers several hours of operation and provide at-home rehabilitation. 

Further work is planned to improve the hybrid actuators by developing alternative exoskeleton and soft actuator design that can be optimized to produce high performances. Folded aluminum sheet can be used as an alternative material to reduce the size and weight while providing more strength. The mix of Ecoflex and Dragonskin can be further tested to find the optimal ratio or a different soft material altogether.

\addtolength{\textheight}{-2cm}   




\bibliographystyle{IEEEtran}
\bibliography{root}

\end{document}